%% file: 1_MainText.tex
\def\year{2022}\relax
\documentclass[letterpaper]{article} % DO NOT CHANGE THIS
\usepackage{aaai22}  % DO NOT CHANGE THIS
\usepackage{times}  % DO NOT CHANGE THIS
\usepackage{helvet}  % DO NOT CHANGE THIS
\usepackage{courier}  % DO NOT CHANGE THIS
\usepackage[hyphens]{url}  % DO NOT CHANGE THIS
\usepackage{graphicx} % DO NOT CHANGE THIS
\usepackage{natbib}  % DO NOT CHANGE THIS AND DO NOT ADD ANY OPTIONS TO IT
\usepackage{caption} % DO NOT CHANGE THIS AND DO NOT ADD ANY OPTIONS TO IT
\DeclareCaptionStyle{ruled}{labelfont=normalfont,labelsep=colon,strut=off} % DO NOT CHANGE THIS
\usepackage{algorithm}
\usepackage{algorithmic}

\usepackage{subfigure}
\usepackage{booktabs} % for professional tables
\usepackage{makecell}

\usepackage{amsmath,amsfonts,amssymb,amsthm}
\usepackage{theoremref,color}
\usepackage{multicol,multirow}
\usepackage{setspace}
\usepackage{algorithm}
\usepackage{algorithmic}
\usepackage{graphicx}
\usepackage{pifont}
\usepackage{makecell}
\usepackage{wrapfig} 
\usepackage{xcolor}
\usepackage{bbm}
\usepackage{dsfont}
\usepackage{amssymb}

\allowdisplaybreaks

\newtheorem{Theorem}{Theorem}
\newtheorem{Proposition}{Proposition}

\theoremstyle{remark}

\DeclareMathOperator*{\argmin}{arg\,min}

\newcommand{\EnvSet}{\mathcal{E}}

\newcommand{\EnvFun}[2]{#1_e(#2)}

\newcommand{\TrueFun}[2]{#1_*(#2)}
\newcommand{\Fun}[2]{#1(#2)}
\newcommand{\Emp}[1]{\hat{#1}}
\newcommand{\Noise}{\epsilon}

\newcommand{\intZO}{\int_{0}^{1}}

\newcommand{\Rom}[1]{\uppercase\expandafter{\romannumeral#1}}

\newcommand{\ab}[1]{\emph{Ablation Study #1:}}

\usepackage{newfloat}
\usepackage{listings}
\floatstyle{ruled}
\newfloat{listing}{tb}{lst}{}
\floatname{listing}{Listing}
\title{Regularization Penalty Optimization for Addressing Data Quality Variance in OoD Algorithms}

\author {
    Runpeng Yu,\textsuperscript{\rm 1,}\equalcontrib
    Hong Zhu, \textsuperscript{\rm 2,}\equalcontrib
    Kaican Li, \textsuperscript{\rm 2}
    Lanqing Hong,\textsuperscript{\rm 2}
    Rui Zhang, \textsuperscript{\rm 3,}\thanks{Corresponding author.}
    Nanyang Ye,\textsuperscript{\rm 4,}\footnotemark[2]
    Shao-Lun Huang, \textsuperscript{\rm 1}
    Xiuqiang He \textsuperscript{\rm 2}
}
\affiliations {
    \textsuperscript{\rm 1} Tsinghua-Berkeley Shenzhen Institute, Tsinghua University\\
    \textsuperscript{\rm 2} Huawei Noah's Ark Lab\\
    \textsuperscript{\rm 3} www.ruizhang.info\\
    \textsuperscript{\rm 4} Shanghai Jiao Tong University\\
    yrp19@mails.tsinghua.edu.cn, zhuhong8@huawei.com, mjust.lkc@gmail.com, honglanqin@huawei.com, rayteam@yeah.net,  ynylincoln@sjtu.edu.cn, shaolun.huang@sz.tsinghua.edu.cn, hexiuqiang1@huawei.com
}

\begin{document}
\include{2_MainText}
\end{document}

%% file: 2_MainText.tex
% !TeX root = 1_MainText.tex
\maketitle

\begin{abstract}
	Due to the poor generalization performance of traditional empirical risk minimization (ERM) in the case of distributional shift, Out-of-Distribution (OoD) generalization algorithms receive increasing attention.  However, OoD generalization algorithms overlook the great variance in the quality of training data, which significantly compromises the accuracy of these methods. In this paper, we theoretically reveal the relationship between training data quality and algorithm performance and analyze the optimal regularization scheme for Lipschitz regularized invariant risk minimization.  A novel algorithm is proposed based on the theoretical results to alleviate the influence of low-quality data at both the sample level and the domain level. The experiments on both the regression and classification benchmarks validate the effectiveness of our method with statistical significance. 
\end{abstract}
\section{Introduction} \label{sec-intro}
Traditional empirical risk minimization~\cite{vapnik1998statistical} focuses on in-distribution generalization under the assumption that training and testing data are independent and identically distributed (i.i.d.). In practice, however, it is common that the distribution of the training and testing data differ greatly, which has motivated the research of Out-of-Distribution (OoD) generalization~\cite{domainbed,Arjovsky20,cdann,mixup,arjovsky2019invariant}. 
In this work, we identify a common but often overlooked problem under the setting of OoD generalization -- the varying quality of training data among samples (or instances) and across \emph{domains}\footnote{A domain is a data source or a data group via artificial data segmentation based on prior knowledge of data distribution~\cite{groupDRO,arjovsky2019invariant}} -- which could severely hurt model performance in some cases.
We focus on two types of quality of training data: the density of data and the noise (or error\footnote{In the rest of the paper, we omit ``error" and error is implied when we say ``noise".}) rate of data. Training data in the ranges of low density or with high noise rate are deemed as low-quality  data. The variance in data quality mainly exists at two levels, the sample level and the domain level, and we need to address the problem at both levels.

Our intuition to address the data quality problem is that, when applying any OoD generalization algorithm, low-quality  data should be detected and regularized (penalized) differently from normal-quality data. The main challenge is how to properly regularize each sample and domain since the numbers of samples and domains are usually quite large. We base our method on one of the most influential OoD generalization paradigms: invariant risk minimization (IRM)~\cite{arjovsky2019invariant}.
To overcome the aforementioned challenge, we propose Lipschitz regularized IRM (LipIRM) loss with varied regularization coefficients on different samples and domains. We then theoretically reveal the links among algorithm performance, data quality (in terms of the density of the data and variance of the noise) and regularization. Based on the theoretical analysis, we propose a regularization scheme that is optimal in the sense that it minimizes the distance between the estimated model and the ground truth model (measured by Eq.~\ref{metric}).

Our contributions are summarized as follows: (i) We identify the performance degeneration of the OoD generalization algorithms when trained on data of varied quality, and we provide the first solution to this problem. (ii) We theoretically derive the optimal regularization scheme for IRM with the Lipschitz regularizer. (iii) Based on the theoretical analysis, we propose a novel algorithm named \textbf{R}egularization \textbf{P}enalty \textbf{O}ptimization (RPO) to address the data quality variance problem, which is via sample-wise and domain-wise regularization. (iv) We conduct extensive experiments on publicly available real-world datasets, and the results show that RPO outperforms the state-of-the-art algorithms.

\section{Preliminaries}
\paragraph{OoD Generalization.} 
In the setting of OoD generalization, a predictor (e.g., neural network) is trained on multiple training datasets generated from different domains and is evaluated on unseen testing domains~\cite{domainbed}. Let $\EnvSet_{all}$ and $\EnvSet_{tr} \subset \EnvSet_{all}$ denote the set of all possible domains and the set of the domains generating the available training datasets, respectively. Let $D^e \triangleq \{(x_e^{(i)},y_e^{(i)})\}_{i=1}^{N_e}$ denotes the dataset generated from the domain $e$, where $e\in \EnvSet_{all}$, $x_e^{(i)}\in\mathcal{X}$ is the features of the $i$-th sample, $y_e^{(i)}\in\mathcal{Y}$ is the corresponding label and $N_e$ is the sample size of domain $e$. 
Given the training datasets $D_{tr} \triangleq \{D^e \mid e\in\EnvSet_{tr}\}$, 
the goal of the OoD problem is to minimize the maximum risk across all the domains, namely to minimize $\max_{e\in\mathcal{E}_{all}}R^e(f)$, where $f\in \mathcal{F}:\mathcal{X}\rightarrow\mathcal{Y} $ is the predictor, and $R^e(f)$ is the risk evaluated in domain $e$.

\paragraph{IRM.} Under the framework of IRM, the predictor $f$ is decomposed into a concatenation of a representation function $\Phi:\mathcal{X}\rightarrow\mathcal{H}$ and a classifier $w:\mathcal{H}\rightarrow\mathcal{Y}$, i.e. $f = w\circ\Phi$. To guarantee the existence of the unified optimal predictor over $\EnvSet_{all}$, IRM utilizes the invariant condition ~\cite{ahuja2021empirical}, which assumes that there exists a data representation $\Phi_{*}$ inducing an invariant predictor $w_*\circ\Phi_{*}$, where the classifier $w_*$ is optimal over all training domains, simultaneously (i.e. $\forall e\in\mathcal{E}_{tr}$, $w_* \in \argmin_{\bar{w}:\mathcal{H}\rightarrow\mathcal{Y}} R^e(\bar{w}\circ\Phi_{*})$). It searches for such invariant predictor $w\circ\Phi$ by minimizing the empirical risk with an extra constraint on the unified optimality of $w$, that is
\begin{gather*}
	\min_{\substack{\Phi:\mathcal{X}\rightarrow\mathcal{H}\\ w:\mathcal{H}\rightarrow\mathcal{Y}}} \sum_{e\in\EnvSet_{tr}} R^e(w\circ\Phi),\\ \quad
	s.t.\quad w \in \argmin_{\bar{w}:\mathcal{H}\rightarrow\mathcal{Y}} R^e(\bar{w}\circ\Phi_{*}), \forall e\in\mathcal{E}_{tr}.
\end{gather*}

In order to solve the above bi-leveled optimization problem, IRM utilizes the augmented Lagrangian method to transfer the constraint to a gradient regularization on each domain $e$. At last, the object function of IRM is expressed as 
\begin{align*}
	\min_{\substack{\Phi:\mathcal{X}\rightarrow\mathcal{H}\\ w:\mathcal{H}\rightarrow\mathcal{Y}}} \sum_{e\in\EnvSet_{tr}} R^e(\Phi) +\eta \lVert\nabla_{w|w=1.0}R^e(w\Phi)\rVert_2^2,
\end{align*}
where the classifier $w$ reduces to a scalar and fixed ``dummy'' classifier; $\Phi$ becomes the entire invariant predictor; $\eta$ is the unified IRM regularization coefficient, which, later in our setting, will vary across domains and serve as the tool of downweighting the domains with low-quality  data.	

\paragraph{Lipschitz regularization.} 
Lipschitz regularization is first used in the statistical regression problems~\cite{wang2013smoothing} and recently proved to have a better generalization guarantee~\cite{wei2019data,wei2019improved} and be a sufficient condition for the smoothness of the representation function~\cite{lipsmooth} in the deep learning context. It penalizes the gradient of the output of the model corresponding to input features. 
Moreover, the strength of the penalties can be designed individually for each sample. So, we can assign large Lipschitz regularization penalties to the low-quality  data to prevent the model from fitting on them. By doing this, the negative influence of the low-quality  data is mitigated by the Lipschitz regularization. Comparatively, the techniques which punish uniformly on the entire model, such as $l_n$ norm regularization, cannot filter out the low-quality  data and thus lacks the robustness against the low-quality  data.

\section{Our Method} \label{Our Method}
\begin{table}[t]
	\centering
	\begin{tabular}{@{}p{1cm}<{\centering}p{7cm}<{\raggedright}@{}}
		\toprule
		Symbol               & Description \\\midrule        	
		$f_*(x)$               & The ground true function projecting from $\mathcal{X}$ to $\mathcal{Y}$.           \\
		$\hat{f}(x)$           & The estimation of the ground true function.           \\
		$\mathcal{R}(\hat{f})$ & The expected MSE of the $\hat{f}$ used to evaluate the goodness of $\hat{f}$.           \\
		$\eta_e$               & The penalty coefficient before the IRM regularization of domain $e$.           \\
		$\lambda$              & The scale of the penalty coefficient before the Lipschitz regularization. \\
		$\rho(x)$              & The relative value of penalty coefficient before the Lipschitz regularization for input feature $x$.          \\
		$\hat{r}_e(x)$         & The empirical probability density of input feature  $x$ in domain $e$.           \\
		$r_e(x)$               & The true probability density of %input 
		$x$ in domain $e$.           \\
		$r(x)$                 & The summation of the true density of $x$ over training domains.          \\
		$\sigma_e^2(x)$        & The variance of the noise of input  $x$ in domain $e$.           \\
		$\hat{r}_{e,k}$        & The groupwise mean of $\hat{r}_e(x)$ in the $k$-th group.           \\
		$\sigma_{e,k}^2$       & The groupwise mean of $\sigma_e^2(x)$ in the $k$-th group.           \\
		$N_e$                  & The sample size of domain $e$.          \\ \bottomrule
	\end{tabular}
	\caption{Frequently Used Symbols}
	\label{tab:importantnotation}
\end{table}

As discussed in Section~\ref{sec-intro}, data quality variance exists at two levels, the sample-level and the domain-level, so we propose a Lipschitz regularized IRM (LipIRM) loss to realize the different regularization at each level. Our key insight is that the second term of the IRM objective function may be regarded as a domain-level regularization to the ERM loss, and Lipschitz regularization may further add different sample-level penalties to different samples. Formally, our proposed LipIRM loss is
\begin{align}
	\label{loss}
	L(f, X, Y) = &\underbrace{\sum_{e\in\EnvSet_{tr}} R^e(\Phi)}_{\text{ERM}}
	+\underbrace{\sum_{e\in\EnvSet_{tr}}\eta_e \lVert\nabla_{w|w=1.0}R^e(w\Phi)\rVert_2^2}_{\text{IRM regularization}} \nonumber\\
	&+\underbrace{\lambda\int_{\mathcal{X}}\rho(x)[f'(x)]^2dx}_{\text{Lipschitz regularization}},
\end{align}
where $\lambda$ is a hyperparameter for the Lipschitz regularization on the whole set of samples, $\eta_e$ is a hyperparameter for the IRM regularization in domain $e$, and $\rho(x)$ is a hyperparameter for the Lipschitz regularization on each sample $x$.
The LipIRM loss consists of three terms, the ERM term, the IRM regularization term (the above-mentioned domain-level regularization), and the Lipschitz regularization term (the above-mentioned sample-level regularization). We call these two regularization terms collectively the \textit{LipIRM regularization}. Note that, the larger the $\eta_e$, the more the model fits the training samples in domain $e$; in contrast, the larger the $\rho(x)$, the less the model fits the sample $x$. Therefore, if a domain has low data quality, $\eta_e$ should be made small, and if a sample has low quality, $\rho(x)$ should be made large. Unlike the usual approach of grid search to obtain the values for these hyperparameters, we will theoretically derive the optimal values for them below. Table~\ref{tab:importantnotation} summarizes frequently used symbols. When there is no ambiguity, we simplify the notation of function by removing its argument. For example, sometimes we may abbreviate $f(x)$ to $f$.

The estimated optimal model $\hat{f}$ (e.g. a well-trained neural network) is obtained by minimizing the loss in Eq.~\eqref{loss}, i.e.,
\begin{align} \label{optim}
	\hat{f} = \argmin_{f\in\mathcal{F}}\ L(f, X, Y).
\end{align}

We assume there is an underlying ground truth $f_*\in\mathcal{F}$, and for each sample $(x_e^{(i)},y_e^{(i)})$ in domain $e$, label $y_e^{(i)}$ equals $f_*(x_e^{(i)})$ plus a relatively small noise $\epsilon_{e}(x_e^{(i)})$, i.e. $y_e^{(i)} = f_*(x_e^{(i)})+\epsilon_{e}(x_e^{(i)})$, and the mean of the noise $\epsilon_{e}(x_e^{(i)})$ is zero. Denote the variance of $\epsilon_{e}(x_e^{(i)})$ by $\sigma_e^2(x_e^{(i)})$, which varies across different domains and samples. To measure the goodness of $\hat{f}$, the expectation of the mean square error (expected MSE for short) is used to calculate the distance between $\hat{f}$ and $f_*$, which is defined as
\begin{align}\label{metric}
	\mathcal{R}(\hat{f}) =\mathbb{E} \big[\int_{\mathcal{X}}[\hat{f}(x)-f_*(x)]^2dx\big],
\end{align}
where the expectation is taken over the noise $\{\epsilon_e \mid e \in \EnvSet_{tr}\}$. Finding the optimal $\hat{f}$ is equivalent to minimizing $\mathcal{R}(\hat{f})$. As our analysis will show later, deriving the optimal solution for generic cases is difficult, because deriving the explicit expression of $\mathcal{R}(\hat{f})$ for high-dimensional function $\hat{f}$ involves solving an intractable system of partial differential equations. Therefore, we first focus our theoretical analysis on a simple case of	one-dimensional regression problem, where $\mathcal{X}=[0,1]$ and $R^e(f)$ is the mean square loss. Under this setting, with mild assumptions, the analytical form of $\mathcal{R}(\hat{f})$ can be obtained as given by \thref{thm:1} below. We discuss in the end of Section~\ref{optimal} how our method may be extended to more generic cases, and our extensive experimental study validates the effectiveness of our method in generic cases.

\subsection{Closed-Form Relationship Among $\mathcal{R}(\hat{f})$, Data Quality and LipIRM Regularization}\label{sec:closedform}
We show in \thref{thm:1} that in the simple case of one-dimensional regression problem,  $\mathcal{R}(\hat{f})$ can be expressed as a closed-form formula of the quality of the training data and the LipIRM regularization. Since the training data are given and the statistics representing the data quality can be estimated from the training data, the theorem can guide us to adjust the regularization hyperparameters according to the estimated data quality so that $\mathcal{R}(\hat{f})$ is minimized. 

Before presenting the theorem, we introduce some notations. Let $\delta(\cdot)$ denote the Dirac delta function. For each domain $e\in\EnvSet_{tr}$, let 
function $\hat{r}_e(x)\triangleq \frac{1}{N_e}\sum_{i=1}^{N_e}\delta(x-x_e^{(i)})$ represents the empirical density of $x$ in domain $e$, and function $r_e(x)$ represent the true density of $x$ in domain $e$. Let $r(x) \triangleq \sum_{e\in\EnvSet_{tr}}r_e(x)$ denote the summation of the true density at $x$.

Further, we make the following assumptions: (i) $\rho(x)$ is chosen to satisfy that $\forall x \in [0,1]$, $\rho(x)$ is positive, and $\rho(x)$ is an analytic function; (ii) $\forall x \in [0,1]$, $r(x)$ is positive and analytic; (iii) $f_*$ is third-order differentiable, which usually holds in mainstream models; (iv) $\lambda \ll 1$, which holds since the regularization term usually has a weight less than 0.1; (v) the noise $\Noise_{e}(x)$ is small compared with $f_*(x)$. That is, let us decompose $\Noise_{e}(x)$ as $\Noise_{e}(x) = \EnvFun{\tau}{x}\upsilon$, where $\upsilon$ is a random variable with $\mathbb{E}[\upsilon] = 0$ and $\mathrm{Var}[\upsilon] = 1$, $\EnvFun{\tau}{x}$ is a deterministic function. Then $\EnvFun{\tau}{x}$ should satisfy $\sup_{x\in [0,1]}\left|\frac{\EnvFun{\tau}{x}}{f_*(x)}\right|\ll 1$.

\begin{Theorem}[Evaluation of the expected MSE $\mathcal{R}(\hat{f})$]\textcolor{white}{s}\thlabel{thm:1}
	\begin{equation*}
		\begin{aligned}
			\Fun{\mathcal{R}}{\Emp{f}} = \intZO\big[\mathbb{E}^2[\Fun{\Emp{f}}{t}-\TrueFun{f}{t}]+\mathrm{Var}[\Fun{\Emp{f}}{t}-\TrueFun{f}{t}]\big]dt,
		\end{aligned}
	\end{equation*}
	where
	\begin{align*}
		&\mathbb{E}^2[\hat{f}(x)-f_*(x)] =\\ &\qquad\lambda^2\left[\frac{[\rho(x)f_*'(x)]'}{r(x)}-\sum_{e\in\EnvSet_{tr}} \frac{\hat{r}_e(x)f_*(x)}{4\eta_{e}r(x)\intZO \frac{\hat{r}_{e}^3(t)}{r^2(t)}f_*^2(t)dt}\right]^2,\\ &\mathrm{Var}[\hat{f}(x)-f_*(x)] =\frac{1}{\sqrt{\lambda}}\sum_{e\in\EnvSet_{tr}} \frac{\hat{r}_e(x)\sigma_{e}^2(x)}{N_er(x)\sqrt{r(x)\rho(x)}}.
	\end{align*}
\end{Theorem}
The crux of proving \thref{thm:1} is the evaluation of $\hat{f}$, which is achieved as follows. Eq.~\eqref{optim} formulates an optimization problem on the functional space, without limiting the analysis to a specific parameterized model. Based on the functional analysis, we derive the sufficient and necessary condition for $\hat{f}$, which is an equation including high order derivative of $\hat{f}$. We then convert the evaluation of $\hat{f}$ to a boundary value problem, which largely involves solving complicated differential equations. To solve these complicated partial differential equations, we utilize the technique of Green functions~\cite{greenf}. After obtaining $\hat{f}$, substituting it into Eq.~\eqref{metric} completes the proof. See Appendix~B.1-B.6 for the detailed derivation.

Based on \thref{thm:1}, we derive the LipIRM regularization scheme that minimizes $\mathcal{R}(\hat{f})$, which we call \textit{the optimal LipIRM regularization scheme}, in Section~\ref{optimal}.

\subsection{Optimal LipIRM Regularization}\label{optimal}

We derive the values of the hyperparameters $\lambda$, $\eta_{e}$, and $\rho$ for the optimal LipIRM regularization. First, we find the optimal value for $\lambda$ by computing the partial derivative of $\mathcal{R}(\hat{f})$ in \thref{thm:1} over $\lambda$. We obtain that $\lambda= (\sum_{e\in\EnvSet_{tr}}\frac{1}{N_e})^{\frac{2}{5}}$. See details in Appendix B.7.

Next, we find optimal values of $\rho$ and $\eta_{e}$. Instead of finding the optimal continuous function $\rho^*$, we parameterize $\rho$ as a group-wise constant function, then find the optimal group-wise $\rho$. More specifically, suppose the support of $x$ can be divided into $K$ small groups, $[x_0,x_1),[x_1,x_2),\cdots,[x_{K-1},x_K)$. For the $k$-th group, $k=1,\cdots,K$, we use the constant $\rho_k$ as the Lipschitz regularization penalty for all of the $x$ in this group. We further constrain that the data samples belonging to the same group are in the same domain. It turns out that the optimal $\eta_{e}$ and $\rho$ are related to the variance of the noise $\sigma_e^2(x)$ and the empirical density $\hat{r}_e(x)$, for $x\in\mathcal{X}$ and $e\in\EnvSet_{tr}$. Instead of estimating $\sigma_e^2(x)$ and $\hat{r}_e(x)$ for each $x\in\mathcal{X}$ and $e\in\EnvSet_{tr}$, we also use the group-wise average to approximate them. For all of the samples in domain $e$ and the $k$-th group, let $\sigma_{e,k}^2$ denote their average variance of the noise, and let $\hat{r}_{e,k}$ denote their average empirical density. The following proposition provides the estimation of the optimal $\eta_{e}$ and $\rho$. The main idea of the proof is to let the derivative of $\mathcal{R}(\hat{f})$ be zero corresponding to $\eta_{e}$ and $\rho$. See Appendix~B.8 for the details.
\begin{Proposition}\thlabel{prp:1}
	For domain $e\in\EnvSet_{tr}$ and group $k=1,\cdots,K$, the optimal $\rho_k^*$ and optimal $\eta_e^*$ are
	\begin{equation*}
		\begin{split}
			\rho_k^* &= \frac{1}{4^{\frac{2}{5}}}\sum_{e\in\EnvSet_{tr}}\left(\frac{ \sigma_{e,k}}{\hat{r}_{e,k}}\right)^{\frac{4}{5}}\mathds{1}_{e,k},
			\\
			\eta_e^* &=\frac{N_e}{4^{\frac{7}{5}}} \left[\sum_{k=1}^{K}\left(\frac{ \sigma_{e,k}}{\hat{r}_{e,k}}\right)^{\frac{4}{5}}\mathds{1}_{e,k}\right]^{-1},
		\end{split}
	\end{equation*}
	where the indicator $\mathds{1}_{e,k}=1$ if the $k$-th group contains the data samples in domain $e$, and $\mathds{1}_{e,k}=0$ otherwise.
\end{Proposition} 

The optimal $\rho_k^*$ is proportional to $\sigma_{e,k}^{\frac{4}{5}}$ and $\hat{r}_{e,k}^{-\frac{4}{5}}$, which indicates that we should fit more on the data samples that have high density %\textcolor{blue}{sufficient data samples} 
and less noise. The optimal $\eta_{e}^*$ is proportional to $N_e$, which indicates that LipIRM should fit more on the domain with sufficient data samples. %of \textcolor{blue}{high density}
Another factor in $\eta_e^*$ is the reciprocal of the weighted sum of $\hat{r}_{e,k}^{-\frac{4}{5}}$, whose weight is an exponent of the variance of the noise $\sigma_{e,k}^{\frac{4}{5}}$. Then we have the following observations for $\eta_e^*$: (i) In domain $e$, if the variance of the noise is a constant, the more density variance $e$ has, the smaller $\eta_{e}^*$ is; (ii) In domain $e$, the greater the variance of the noise is,  the smaller $\eta_{e}^*$ is; (iii) In domain $e$, if the groups of greater noise have smaller density, $\eta_{e}^*$ will be small. In summary, we infer that LipIRM should emphasize fitting the domains that have small density variance and low noise rate. 

\paragraph{Extension to Generic Cases.}

So far, our analysis and derivations have been based on the simple case of one-dimensional regression problem, where $\mathcal{X}=[0,1]$ and $R^e(f)$ is the mean square loss. For more generic cases of high-dimensional data or classification tasks, we may not get nice closed-form expressions like \thref{prp:1}, but the insights provided by \thref{prp:1} still hold, i.e., for training data of low quality, $\eta_e$ should be set small and $\rho_k$ should be set large. Thus the expressions in \thref{prp:1} may still be reasonable choices of the optimal penalties. In the experimental study, all the datasets are of high-dimensional data. We have used the method in Section~\ref{optimal} for setting the values of the hyperparameters, and our algorithm performs very well, outperforming state-of-the-art algorithms significantly. The experiments in Section~\ref{exp:cmnist} are for classification tasks, and our algorithm also performs very well. These validate the effectiveness of our method for generic cases.

Finally, the empirical LipIRM loss is written as
\begin{align} \label{optimal-loss}
	\hat{L}(f,D_{tr}) &= \sum_{e\in\EnvSet_{tr}}\sum_{i=1}^{N_e}l(f(x_e^{(i)}),y_e^{(i)})\nonumber\\&+\sum_{e\in\EnvSet_{tr}}\eta_e^* \lVert\nabla_{w|w=1.0}[\sum_{i=1}^{N_e}l(wf(x_e^{(i)}),y_e^{(i)})]\rVert_2^2\nonumber\\&+\lambda\sum_{e\in\EnvSet}\sum_{i=1}^{N_e}\rho^*(x_e^{(i)})[f'(x_e^{(i)})]^2,
\end{align}
where $l(\cdot,\cdot)$ is the loss function over a sample, $\rho_k^*$ and optimal $\eta_e^*$ are the optimal penalties obtained in \thref{prp:1}, $\rho^*(x_e^{(i)})=\rho_k^*$ if sample $(x_e^{(i)},y_e^{(i)})$ is in group $k$.

\paragraph{Discussions on Implementation.}
There are two steps left before using the expressions in \thref{prp:1}: (i) grouping the data and (ii) estimating the statistics ($\hat{r}_{e,k}$ and $\sigma_{e,k}^2$). Note that we should not group the data arbitrarily. For example, the naive way of evenly dividing after randomly shuffling the data will make all the groups have similar frequencies and noise rates, and hence fails to reflect the data quality variance in the original data. The right way to group the data should guarantee that the samples in the same group are close in the feature space and possess similar  statistical  properties. This kind of grouping can be treated as a clustering problem with constraints on the statistical similarity of the data in each cluster, and we can solve it by the axis parallel subspace clustering algorithm~\cite{highcluster}. Specifically, we first select several dominating features which can be regarded as the main causes of the  statistical heterogeneity and then cluster the samples in the subspace of these features.
The dominated features can be identified by human experts if the dataset has explicit features. In the task of image classification where there are no explicit features, the class label can be used as the dominated feature, e.g. the samples with the same digit in the Colored-MNIST or the same type of the animal (or vehicle) in NICO are grouped into the same group. 

For estimating the statistics, (i) $\hat{r}_{e,k}$ can be estimated from the empirical density (i.e. $\hat{r}_{e,k}=\frac{N_{e,k}}{N_e}$, where $N_{e,k}$ is the number of samples in domain $e$ and the $k$-th group). (ii) 
Because the variance can be rewritten as $\sigma^2_{e}(x) =\mathbb{E}\big[[\epsilon_{e}(x)]^2\big]=\mathbb{E}\big[[y_e(x)-f_*(x)]^2\big]$, where $y_e(x)=f_*(x)+\epsilon_{e}(x)$ is the observed label, and the expectation is taken over the random noise $\epsilon_{e}(x)$. $f_*(x)$ is indeed unknown, so we propose to replace it with the estimation $\tilde{f}(x)$ and approximate the variance via $\sigma^2_{e}(x) \approx  \mathbb{E}\big[[y_e(x)-\tilde{f}(x)]^2\big]$. Therefore, we first train an extra model $\tilde{f}$ and then use the average prediction error of $\tilde{f}$ over the samples in the domain $e$ and group $k$ to approximate $\sigma^2_{e,k}$. Our algorithm is summarized in Algorithm \ref{alg:1}.

\begin{algorithm}[tb]
	\caption{Regularization Penalty Optimization (RPO)}
	\label{alg:1}
	\begin{algorithmic}[1]
		\REQUIRE The training datasets $D_{tr} = \{D^e \mid e\in\EnvSet_{tr}\}$ and a parameterized model $f_{\mathbf{\theta}}$ where $\mathbf{\theta}$ is the parameters.
		\STATE Randomly initialize the model parameters: $\mathbf{\theta}\gets \mathbf{\theta}_0'$.
		\STATE For all $e$ and $k$, $\rho_k \gets 1$, $\eta_e \gets 1$
		\STATE $f_{\mathbf{\theta}_*'} \gets$ SGD with LipIRM loss in Eq.~\eqref{optimal-loss} on $D_{tr}$
		\STATE For all $e$ and $k$, estimate $\hat{r}_{e,k}$, $\sigma_{e,k}^2$
		\STATE For all $e$ and $k$, compute  $\rho_k^*$ and $\eta_e^*$ using \thref{prp:1}
		\STATE Randomly reinitialize the model parameters: $\mathbf{\theta}\gets \mathbf{\theta}_0$.
		\STATE $f_{\mathbf{\theta}_*} \gets$ SGD with LipIRM loss in Eq.~\eqref{optimal-loss} on $D_{tr}$
	\end{algorithmic}
\end{algorithm}

\section{Experiments}\label{exp}
We experimentally evaluate the performance of our proposed algorithm RPO on four datasets, two (Cigar and Wage) for regression and two (Colored-MNIST and NICO) for classification. 
Following existing work on these tasks, all the results are averaged over multiple runs under the same setup with different random seeds. For each run, $\rho_i$ and $\eta_e$ in RPO are calculated according to Algorithm \ref{alg:1}. Other hyperparameters in RPO such as learning rates and batch size and the hyperparameters in baselines are optimized by grid research on an independently divided validation set.

\subsection{Baselines}
We compared with a large number of baselines and listed blow. The details on them are provided in Appendix~C.1. We evaluate all the 19 baselines for classification tasks. However, because most of the baselines are designed only for classification, we evaluate those that can also perform regression (6 of them) for regression tasks.

\paragraph{ERM.} For completeness, we add the ERM algorithm into our baselines to indicate the rationality of the OoD method and the Lipschitz regularizer. We implement two types of ERM loss, one with a $l_2$-norm regularizer (ERM + $l_2$), and the other with a uniform Lipschitz regularizer (ERM + $lip$). 

\paragraph{OoD generalization methods.} We compare our methods with the following OoD generalization methods: ANDMask~\cite{andmask}, CDANN~\cite{cdann}, CORAL~\cite{coral}, DANN~\cite{dann}, GroupDRO~\cite{groupDRO}, IGA~\cite{iga}, and the original IRM~\cite{arjovsky2019invariant} with either a uniform $l_2$-norm regularizer (IRM + $l_2$) or a uniform Lipschitz regularizer (IRM + $lip$), Mixup~\cite{mixup}, MLDG~\cite{mldg}, MTL~\cite{mtl}, RSC~\cite{rsc}, SagNet~\cite{sagnet}, and SD~\cite{gs}.

\paragraph{Combination of IRM and methods for learning from imbalanced or noisy data.} In the context of in-distribution generalization, many algorithms have been proposed to promote the performance of learning from the imbalanced~\cite{Cao19,Liu19} or noisy  data~\cite{LiWZK19,XuCKW19,ChenLCZ19,Huang20}. We select three recently proposed algorithms from them and combine IRM with each of them to create three new baselines. We compare our methods with these three baselines to show that, when learning from imbalanced or noisy data, simply adopting the ideas in the algorithms built for in-distribution generalization does not work well for OoD problems. 
The algorithms we choose are:
an imbalanced learning algorithm, LDAM-DRW~\cite{Cao19}, and a noisy label learning algorithm, Self-Adaptive Training (SAT)~\cite{Huang20}. We also compare with HAR~\cite{cao2021heteroskedastic} which addresses the imbalance and label noise in a uniform way by adaptive regularization technique.

\paragraph{Sample reweighting method for OoD problems.} Since the idea of our algorithm is related to sample reweighting methods, we compare with CNBB~\cite{nico} which is the state-of-the-art that uses sample reweighting to address OoD generalization problems.

\subsection{$\boldsymbol{t}$-test} \label{t-test}
\begin{figure*}[t]
	\centering
	\makebox[\textwidth][c]{
		\includegraphics[width=0.9\linewidth]{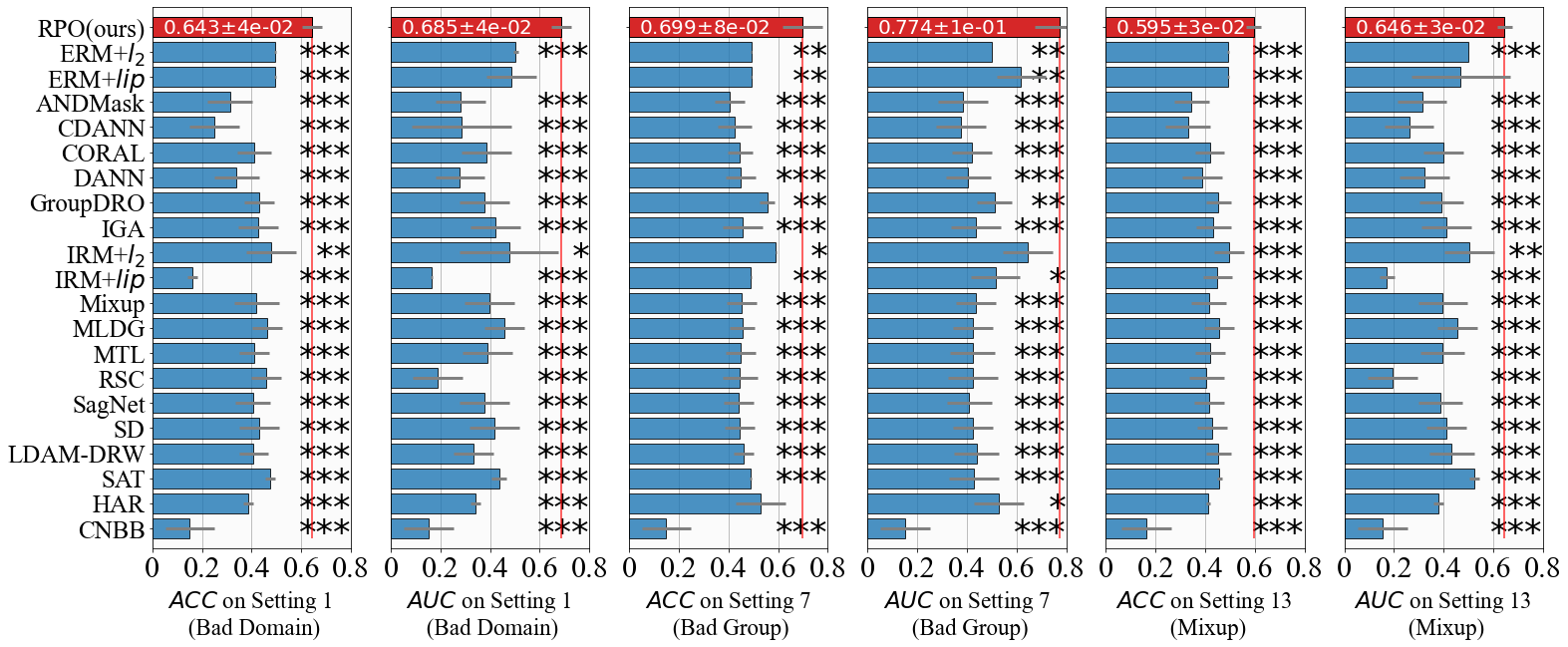}
	}
	\caption{Performances on Colored-MNIST under three kinds of data setting (Bad Domain, Bad Group, Mixup). The horizontal bars indicate the accuracy ($ACC$) or the Area under the ROC Curve ($AUC$) of the method. The narrow gray bars indicate the standard deviation. ``***", ``**" and ``*" indicate the significance of the $t$-test (see Section \ref{t-test}).}
	\label{exp:cmnist:accauc}
\end{figure*}

The $t$-test provides a more strict examination of the performance gap between two methods, so besides reporting the average performance metric (e.g. the mean square error or the mean accuracy), for each baseline, we carry out the Welch's $t$-test of the null hypothesis that the mean of certain performance metric of our method is equal to the mean of that metric of the baseline. The signs ``***",``**" and ``*" in Table \ref{tab:1}, Figure~\ref{exp:cmnist:accauc} and Figure~\ref{exp:nico:accauc} are used to indicate that the $p$-value of the $t$-test is less than $0.01$, $0.05$ and $0.1$, respectively. The more ``*" a baseline has, the more significant performance gain our method has compared to that baseline.

\begin{table}[t]
	\centering
	\begin{tabular}{@{}lr@{ $\pm$ }rr@{ $\pm$ }r@{}}
		\toprule
		& \multicolumn{2}{c}{Cigar}& \multicolumn{2}{c}{Wage}\\ \midrule
		ERM + $l_2$ & 29.85&2.90\tiny$^{***}$  & \ 64.33&13.50\tiny$^{***}$   \\
		ERM + $lip$ & 6.90&0.95\tiny$^{***}$  & 34.95&7.09\tiny$^{***}$   \\
		ANDMask   & 16.63&10.52\tiny$^{***}$ & 101.01&73.38\tiny$^{***}$ \\
		IRM + $l_2$ & 24.79&1.76\tiny$^{***}$   & 49.16&7.94\tiny$^{***}$   \\
		IRM + $lip$ & 6.80&0.84\tiny$^{***}$  & 34.93&7.12\tiny$^{***}$   \\
		MLDG      & 6.54&3.22\tiny$^{***}$  & 114.72&85.14\tiny$^{***}$  \\ \midrule
		%HAR       & 23.87&1.20\tiny$^{***}$  & 39.30&4.95\tiny$^{***}$   \\ 
		RPO (ours) & \textbf{4.35}&\textbf{0.33}\tiny$^{\hphantom{***}}$ & \textbf{2.82}&\textbf{1.43}\tiny$^{\hphantom{***}}$   \\ \bottomrule
	\end{tabular}
	\caption{Performances on regression datasets Cigar and Wage, measured by Mean Square Error (MSE) $\pm$ its standard deviation. ``***", ``**" and ``*" indicate the significance of the $t$-test (see Section \ref{t-test}).} 
	\label{tab:1}
\end{table}

\subsection{Regression Experiments}\label{exp:reg}
Cigar~\cite{BALTAGI1992321,baltagi2021econometric} and Wage~\cite{wooldridge2016introductory} are two commonly used datasets on regression tasks in the study of fixed effect model~\cite{wooldridge2016introductory,rosner2010fundamentals}. These datasets satisfy the assumption that the statistical differences in data are caused by only a few dominating features. We construct different domains by adding different confounders (or in other words, non-causal features) following the procedure in~\cite{arjovsky2019invariant}, where the confounders are generated by different linear transformations of the label plus Gaussian noises. In the training domains, the confounders are closely related to the label (in the sense of Pearson Correlation Coefficient). In the testing domain, the variances of the Gaussian noises are increased to decorrelate the confounders with the label. This also makes the scale of the confounders in the testing domain much larger than that in the training domains. So, in the testing domain, the MSE of the predictor which utilizes the confounders to draw prediction should be much larger than the MSE of the predictor which filters out the confounders. We use this phenomenon to detect whether the OoD generalization algorithms can successfully learn the causal features.  As shown by the results in Table \ref{tab:1}, the MSE of our method is much smaller than other baselines, especially, on the Wage dataset. This indicates that our method has the best OoD performance. Some additional descriptions are shown in Appendix~C.2 to C.4. 

\subsection{Classification Experiments}	\label{exp:cmnist}
Colored-MNIST~\cite{arjovsky2019invariant} and NICO~\cite{nico} are two commonly used datasets for OoD generalization problem on classification tasks.

Colored-MNIST endows the grayscale MNIST image with a binary color attribute to simulate the spurious correlation in the training set. We focus on a harder but more practical setting by introducing density variance and noise rate variance into Colored-MNIST. 

We systematically create 14 experiment settings (see Appendix~ C.5) to simulate various kinds of distributions of the data with density variance and noise rate variance. Here, the results of setting 1 (Bad Domain\footnote{Bad Domain. There are certain low-quality  domains in which a part of the data has high density variance and noise rate variance.}), setting 7 (Bad Group\footnote{Bad Group. Suppose the data can be divided into groups according to some prior knowledge. There are certain low-quality  groups suffering from density variance and noise rate variance.}), and setting 13 (Mixup\footnote{Mixup. The Bad Group and Bad Domain occur together.}) are shown in Figure \ref{exp:cmnist:accauc}. An illustration of setting 1 and setting 7 are shown in Figure \ref{fig:weight}. It is worth noticing that our method achieves statistical significant performance gain over all baselines, except the $AUC$ of ERM+$lip$ on Setting 1, IRM+$l_2$ on Setting 7, and ERM+$lip$ on Setting 13, and the $ACC$ of HAR on Setting 7. We attribute these failure cases to the fact that the performance of these algorithms are unstable. The variances of the $AUC$ (or $ACC$) of these baselines are large so that the $t$-test fails to distinguish our method from them. Despite lack of statistical significance in these three comparisons, our method is stabler (has smaller variance) and achieves better average performance than them. 

\begin{figure}[t]
	
	\centering
	\includegraphics[width=0.85\columnwidth]{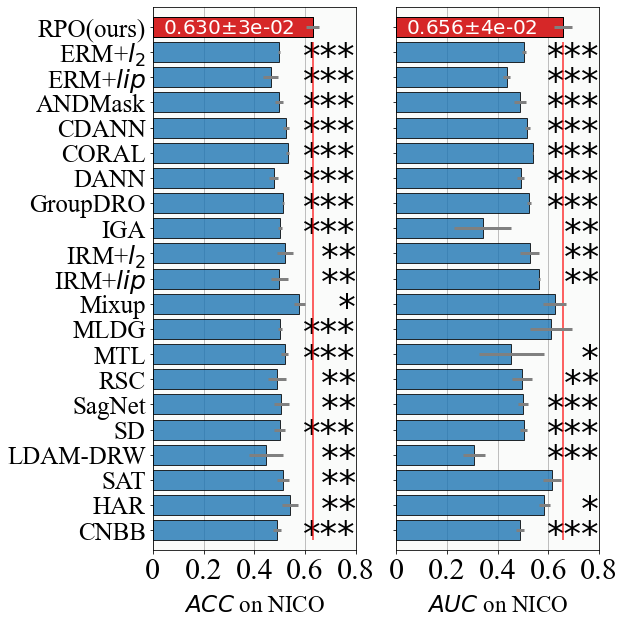}
	\caption{Performances on NICO. The horizontal bars indicate the accuracy ($ACC$) or the Area under the ROC Curve ($AUC$) of the method. The narrow gray bars indicate the standard deviation. ``***", ``**" and ``*" indicate the significance of the $t$-test (see Section \ref{t-test}).} 
	\label{exp:nico:accauc}
\end{figure}

NICO consists of real-world photos of animals and vehicles taken in a large variety of contexts, for example ``on snow", ``on beach" and ``on grass". Our version of NICO is a binary classification problem where the class labels $Animal$ and $Vehicle$ are spuriously correlated to the photo context. Density variance and noise rate variance are stimulated following the same procedure as generating Setting 13 in our experiment of Colored-MNIST. As shown in Figure \ref{exp:nico:accauc}, RPO achieves state-of-the-art (SOTA) performance in terms of both $ACC$ and $AUC$. The performance gains are statistically significant comparing to almost all the baselines. This demonstrate the utility of RPO in the complex real-world dataset. Additional descriptions are shown in Appendix~C.6.

\subsection{Further Discussions}
In this subsection, further experimental results are provided.

\begin{figure}[t]
	\centering
	\subfigure[]{
		\begin{minipage}[b]{0.22\textwidth}
			\includegraphics[width=\textwidth]{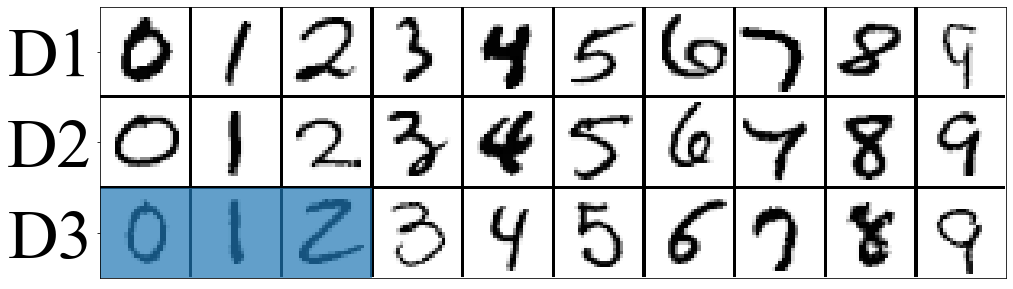}
		\end{minipage}
		\label{fig:weight:a}
	}
	\subfigure[]{
		\begin{minipage}[b]{0.22\textwidth}
			\includegraphics[width=\textwidth]{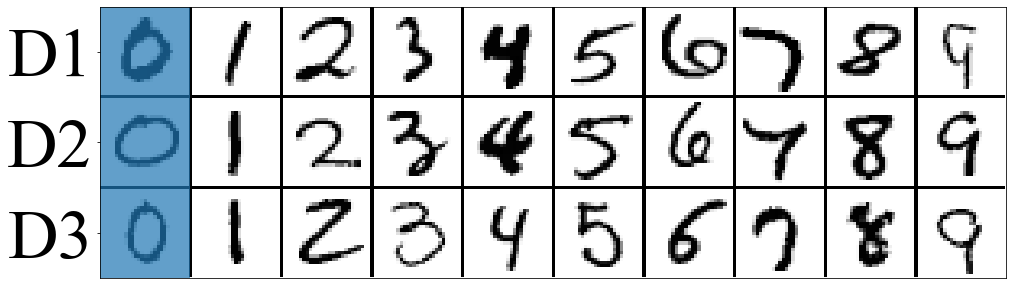}
		\end{minipage}
		\label{fig:weight:b}
	}\\
	\subfigure[]{
		\begin{minipage}[b]{0.22\textwidth}
			\includegraphics[width=\textwidth]{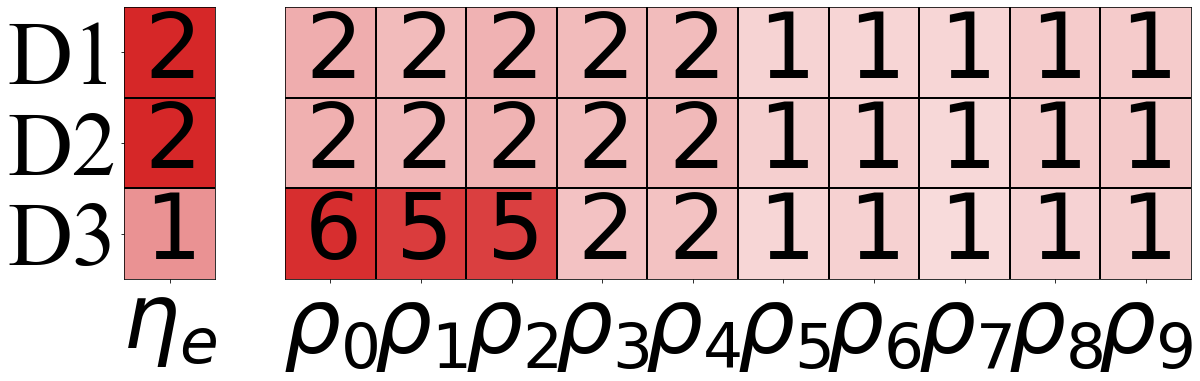}
		\end{minipage}
		\label{fig:weight:c}
	}
	\subfigure[]{
		\begin{minipage}[b]{0.22\textwidth}
			\includegraphics[width=\textwidth]{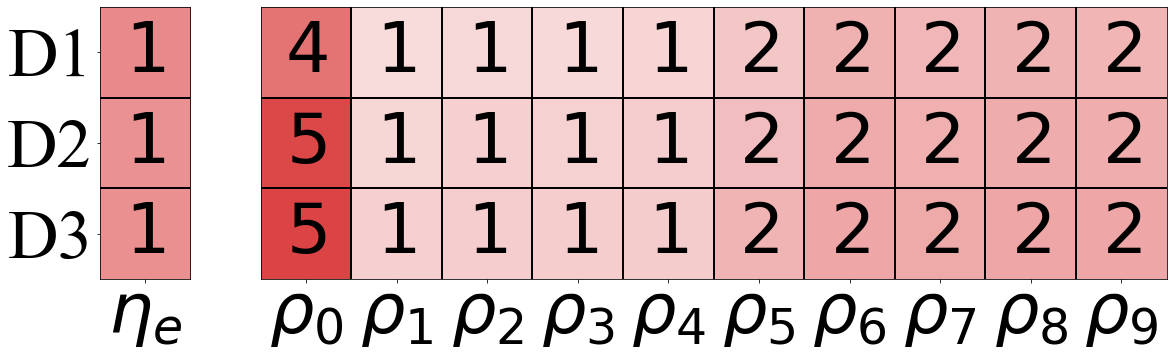}
		\end{minipage}
		\label{fig:weight:d}
	}
	\caption{Subfigure \ref{fig:weight:a} and \ref{fig:weight:b} illustrate our setting 1 and 7, respectively. ``Domain" is abbreviated to ``D" in the figure. We divide the data in each domain into 10 groups according to the digit. We corrupt the data belonging to the groups in blue shadow by downsampling with probability $0.1$ and switching their binary label with probability $0.3$. Though only three groups out of thirty have density variance and noise rate variance, the performance of OoD generalization algorithms declines drastically. Subfigure \ref{fig:weight:c} and \ref{fig:weight:d} present the (relative values of) weights $\eta_{e}$ and $\rho_i$ obtained by our method under setting 1 and 7, respectively.}
	\label{fig:weight}
\end{figure}

\paragraph{Visualization of $\eta_e^*$ and $\rho_i^*$.} A graphical illustration of how we generate setting 1 and 7 together with the regularization penalty obtained by RPO are shown in Figure \ref{fig:weight}. As shown by the figure, our method successfully downweights the low-quality domains with smaller IRM penalties and penalizes the low-quality  data groups with larger Lipschitz regularizations.

\paragraph{Ablation study.}
RPO (our method) uses the optimized penalty for both the IRM term and the Lipschitz regularization term. We compare RPO with \ab{1} RPO-Pen, which uses optimized penalty for the IRM term and uniform regularization for the Lipschitz regularization term, and \ab{2} RPO-Lip,  which uses uniform regularization for the IRM term and optimized penalty for the Lipschitz regularization term. As shown in the Table \ref{tab:4}, for the setting with bad domain, domain-wise reweighting is more effective, and RPO-Pen performs better than RPO-Lip. However, the setting with bad group benefits more from the sample-wise reweighting, and RPO-Lip performs better than RPO-Pen. This is consistent with the intuition that both domain-wise reweighting and sample-wise reweighting are necessary, though, they are suitable for different types of low-quality  data distributions. Moreover, the combination of domain-wise reweighting and sample-wise reweighting leads to further improvements, outperforming RPO-Pen and RPO-Lip in both settings.
\begin{table}[t]
	\small
	\centering
	\begin{tabular}{@{}c@{$\ \ $}c@{$\ \ $}c@{$\ \ $}c@{$\ \ $}c@{}}
		\toprule
		& \multicolumn{2}{c}{\makecell[c]{Setting 1\\(Bad domain)}} & \multicolumn{2}{c}{\makecell[c]{Setting 7\\(Bad Group)}} \\ \midrule
		& ACC           & AUC           & ACC           & AUC           \\
		RPO-Lip & 0.398$\pm$1e-2  & 0.351$\pm$2e-3  & 0.679$\pm$3e-2  & 0.740$\pm$5e-2  \\
		RPO-Pen & 0.616$\pm$5e-2  & 0.649$\pm$7e-2  & 0.544$\pm$1e-1  & 0.555$\pm$1e-1  \\
		RPO     & \textbf{0.643$\pm$4e-2}  & \textbf{0.685$\pm$4e-2}  & \textbf{0.699$\pm$8e-2}  & \textbf{0.774$\pm$1e-1}  \\ \bottomrule
	\end{tabular}
	
	\caption{Ablation study on Colored-MNIST, measured by the accuracy ($ACC$) or the Area under the ROC Curve ($AUC$) $\pm$ corresponding standard deviation.}\label{exp:cmnist:ablation}
	\label{tab:4}
\end{table}

\section{Related Work}
Our work is in the field of OoD generalization. 
Approaches like Bayesian methods~\cite{neal2012bayesian}, data augmentation~\cite{ZhaoFNG19,LiuSHDL20}, robust optimization~\cite{LeeR18,HoffmanMZ18}, and causal-based methods~\cite{kuang2018stable,Rojas-CarullaST18,SubbaswamySS19,sz19} have been proved successful in addressing the OoD generalization problem. In this paper, we take two types of data quality problems into consideration: the density variance and the noise rate variance. In the in-distribution generalization setting, the density variance (also named as imbalance or long-tailed distribution) of data can be alleviated through reweighting~\cite{Cao19,Li0W19}, resampling~\cite{ByrdL19,C0020}, Bayesian learning~\cite{TianLGHK20} and self-supervise learning~\cite{YangX20}. The noise rate variance is new for the deep learning context~\cite{cao2021heteroskedastic} and can be connected to the field of noisy data learning which can be handled by reweighting~\cite{RenZYU18,ShuXY0ZXM19}, self-adaptive learning~\cite{Huang20}, transition matrix estimating~\cite{YaoL0GD0S20} and curriculum learning~\cite{JiangZLLF18}. However, as shown by our experiments, a simple combination of the OoD generalization algorithms with the long-tailed data learning or noisy data learning methods cannot achieve the optimal performance. To recover the performance loss when training with low-quality  data, OoD algorithms need special designs.

\section{Conclusion}
We have identified the problem of poor performance of OoD generalization algorithms caused by the varying quality of training data and provided the first solution to this problem. By theoretically deriving the optimal regularization scheme for IRM with Lipschitz regularizer, we have proposed a novel algorithm named RPO which regularizes via sample-wise and domain-wise penalties. We have conducted extensive experiments on publicly available real-world datasets and the results validate the effectiveness of our method.

\clearpage
\section*{Acknowledgments}
Nanyang Ye was supported by National Natural Science Foundation of China under Grant 62106139. The work of Shao-Lun Huang was supported in part by the National Natural Science Foundation of China under Grant 61807021, in part by the Shenzhen Science and Technology Program under Grant KQTD20170810150821146.
\begin{small}
	\bibliography{ref}
\end{small}